\documentclass[graybox]{svmult}

\usepackage{mathptmx}       
\usepackage{helvet}         
\usepackage{courier}        
\usepackage{type1cm}        
\usepackage{makeidx}         
\usepackage{graphicx}        
\usepackage{multicol}        
\usepackage[bottom]{footmisc}

\usepackage{url}

\usepackage{amsmath}
\usepackage{amssymb}
\usepackage{listings}
\usepackage{kpfonts}
\usepackage{color}
\newcommand*\Hs[1]{\ensuremath{{\color{black} #1}{\color{red}\varheartsuit}}}
\newcommand*\Ss[1]{\ensuremath{{\color{black} #1}{\color{black}\spadesuit}}}
\newcommand*\Ds[1]{\ensuremath{{\color{black} #1}{\color{red}\vardiamondsuit}}}
\newcommand*\Cs[1]{\ensuremath{{\color{black} #1}{\color{black}\clubsuit}}}
\newcommand*\NT[1]{{\color{black} #1}{\color{black}\textsc{NT}}}

\newenvironment{bridgetable}[0]{\newbox\mnord\newbox\msud
\newbox\mest\newbox\mouest\newbox\centre}{
\setbox\centre=\hbox{
\begin{tabular}{|lcr|}
\hline
&N&\\[10pt]
W&&E\\[10pt]
&S&\\
\hline
\end{tabular}
}
\begin{tabular}{ccc}
&\box\mnord\\
&\\
\box\mouest&\box\centre&\box\mest\\
&\\
&\box\msud
\end{tabular}
}

\newenvironment{nord}[0]{\lrbox{\mnord}\tabular{c*{13}{c@{\ }}}}{\endtabular\endlrbox}
\newenvironment{sud}[0]{\lrbox{\msud}\tabular{c*{13}{c@{\ }}}}{\endtabular\endlrbox}
\newenvironment{est}[0]{\lrbox{\mest}\tabular{c*{13}{c@{\ }}}}{\endtabular\endlrbox}
\newenvironment{ouest}[0]{\lrbox{\mouest}\tabular{c*{13}{c@{\ }}}}{\endtabular\endlrbox}
\newcommand\pique[1]{$\Ss{}$&#1\\}
\newcommand\coeur[1]{$\Hs{}$&#1\\}
\newcommand\carreau[1]{$\Ds{}$&#1\\}
\newcommand\trefle[1]{$\Cs{}$&#1\\}

\usepackage{theorem}
\newtheorem{ex}{Exemple}

\usepackage{appendix}

\makeindex             

\begin{document}
\authorrunning{Véronique Ventos et al}
\titlerunning{Construction and Elicitation of a Black Box  Model in the Game of Bridge}
\title*{Construction and Elicitation of a Black Box  Model in the Game of Bridge}
\author{Véronique Ventos,  Daniel Braun, Colin Deheeger, Jean Pierre Desmoulins, Jean Baptiste Fantun,  Swann Legras, Alexis Rimbaud, Céline Rouveirol, Henry Soldano and Solène Thépaut
}

\institute{Daniel Braun, Colin Deheeger, Jean Pierre Desmoulins, Jean Baptiste Fantun, Swann Legras,
     Alexis Rimbaud, Céline Rouveirol, Henry Soldano, Solène Thépaut and Véronique Ventos \at NukkAI, Paris, France
     \and Henry Soldano and Céline Rouveirol  \at Université Sorbonne Paris-Nord, L.I.P.N UMR-CNRS 7030\\ Villetaneuse, France \and Henry Soldano \at Muséum National d'Histoire Naturelle, I.SY.E.B UMR-CNRS 7205, Paris, France}

\maketitle

\abstract*{We address the problem of building a decision model  for a specific bidding situation in the game of Bridge. We propose  the following multi-step methodology i) Build a set of examples for the decision problem and use simulations  to associate a decision to each example ii)  Use  supervised relational learning  to build  an accurate and readable model iii)  Perform a joint analysis between domain experts  and data scientists to  improve the learning language, including the production by experts of a handmade model iv) Use insights from iii to learn a more readable and accurate model.}

\abstract{We address the problem of building a decision model  for a specific bidding situation in the game of Bridge. We propose  the following multi-step methodology: i) Build a set of examples for the decision problem and use simulations  to associate a decision to each example ii)  Use  supervised relational learning  to build  an accurate and interpretable model iii)  Perform a joint analysis between domain experts  and data scientists to  improve the learning language, including the production by experts of a handmade model iv) Use insights from iii) to learn an almost self-explaining  and accurate model.}


\section{Introduction}
Our goal is to model expert decision processes in Bridge. To do so, we propose a methodology involving human experts, black box decision programs, and relational supervised machine learning systems. The aim is to obtain a global model  for this decision process, that is both expressive and has high  predictive performance. 
Following the success of supervised methods of the deep network family, and a growing pressure from society imposing that automated decision processes be made more transparent,
 a growing number of AI researchers are (re)exploring techniques to interpret, justify, or explain "black box" classifiers (referred to as the Black Box Outcome Explanation Problem \cite{GuidottiMRTGP19}). It is a question of 
 building, a posteriori, explicit models in symbolic languages, most often in the form of rules or decision trees that explain the outcome of the classifier in a format intelligible to an expert. 
Such explicit models extracted by supervised learning can carry expert knowledge \cite{inter-ml}, which has intrinsic value for explainability, pedagogy, and evaluation in terms of ethics or equity (they can help to explain biases linked to the learning system or to the set of training examples).
Learning a global model (i.e. capable of explaining the class of any example)  that is both interpretable and highly accurate has been identified as a difficult problem. Many recent approaches that build interpretable models adopt a simpler two-step method: a first step aims to build a global black box classifier that is as precise as possible, a second step focuses on the generation of a set of local explanations (linear models/rules) to justify the classification of a specific example (see for instance the popular LIME systems \cite{LIME} and ANCHORS \cite{ANCHORS} systems).

 
 We present here a complete methodology for acquiring a global model of a black box classifier as a set of relational rules, both as explicit and as accurate as possible. As we consider a game, i.e. a universe with precise and known rules, we are in the favorable  case where it is possible to generate, on demand, data that i)  is in agreement with the problem specification, and ii) can be labelled with correct decisions through simulations.    The methodology we propose consists of the following elements:
 \begin{enumerate}
 \item Problem modelling with relational representations, data generation and labelling.
 \item Initial investigation of the learning task and learning with relational learners.
 \item Interaction with domain experts  who refine a  learned model to produce a simpler  alternative model, which is more easily understandable for domain users.
 \item Subsequent investigation of the learning task, taking into account the concepts used by experts to produce their  alternative model, and the proposition of a new, and more accurate model.
  \end{enumerate}
 The general idea is to maximally leverage interactions between experts and engineers, with each group building on the analysis of the other.

 We approach the  learning task using relational supervised learning methods from \emph{Inductive Logic Programming} (ILP)   \cite{MuggletonR94}.  The language of these methods, a restriction of first-order logic, allows learning compact rules, understandable by experts in the domain.  The logical framework   allows   the use of a domain vocabulary together with domain knowledge defined in a domain theory, as illustrated in early work on the subject \cite{ilpbridgegamechallenge}.  
 
 The outline of the article is as follows. After a brief introduction to bridge in Section \ref{bridgepb}, we describe in Section \ref{genedonnees} the relational formulation of the target learning problem, and the method for generating and labelling examples.
 We then briefly describe in Section \ref{ilp} the ILP systems used, and the first set of experiments run on the target problem, along with their results (Section \ref{expes}). In Section \ref{analyse}, bridge experts review a learned model's output and build a powerful  alternative model of their own, an analysis of which leads to a refinement of the ILP setup and further model improvements. Future research avenues are outlined in the conclusion.

\section{Problem Addressed}
\label{13bridgepb}

Bridge is played by four players in two competing partnerships, namely, \emph{North} and \emph{South} against \emph{East} and \emph{West}. A classic deck of 52 playing cards is shuffled and then dealt evenly amongst the players ($\frac{52}{4}=13$ cards each). The objective of each side is to maximize a score which depends on:
\begin{itemize}
    \item \textbf{The vulnerability} of each side. A \emph{non-vulnerable} side loses a low score when it does not make it's contract, but earns a low score when it does make it. In contrast, a \emph{vulnerable} side has higher risk and reward.
    \item \textbf{The contract} reached at the conclusion of the \emph{auction} (the first phase of the game). The contract is the commitment of a side to win a minimum of $l_{min} \in \{7, \dots 13\}$ tricks in the \emph{playing phase} (the second phase of the game). The contract can either be in a \emph{Trump} suit (\Cs{}, \Ds{}, \Hs{}, \Ss{}), or \emph{No Trumps} (\NT{}), affecting which suit (if any) is to gain extra privileges in the playing phase. An opponent may \emph{Double} a contract, thus imposing a bigger penalty for failing to make the contract (but also a bigger reward for making it). A contract is denoted by $pS$ (or $pS^X$ if it is Doubled), where $p=l_{min} - 6\in \{1, \dots 7\}$ is the \emph{level} of the contract, and $S \in \{\Cs{},\Ds{},\Hs{},\Ss{},NT\}$ the \emph{trump suit}. The holder of the contract is called the \emph{declarer}, and the partner of the declarer is called the \emph{dummy}.
    \item \textbf{The number of tricks won} by the declaring side during the playing phase. A trick containing a trump card is won by the hand playing the highest trump, whereas a trick not containing a trump card is won by the hand playing the highest card of the suit led.
\end{itemize}
\bigskip
 For more details about the game of bridge, the reader can consult \cite{acbl}. Two concepts are essential for the work presented here:
\begin{itemize}
\item {\bf Auction}: This allows each player (the first being called the \emph{dealer}) the opportunity to disclose coded information about their hand or game plan to their partner\footnotemark.
\footnotetext{The coded information given by a player is decipherable by both their partner and the opponents, so one can only deceive their opponents if they're also willing to deceive their partner. In practice, extreme deception in the auction is rare, but for both strategical and practical reasons, the information shared in the auction is usually far from complete.}
Each player bids in turn, clockwise, using as a language the elements: \emph{Pass}, \emph{Double}, or a contract higher than the previous bid (where $\Cs{} < \Ds{} < \Hs{} < \Ss{} < NT$ at each level). The last bid contract, followed by three Passes, is the one that must be played.

\item {\bf The evaluation of the strength of a hand}: Bridge players assign a value for the highest cards: an Ace is worth 4\,HCP (\emph{High Card Points}), a King 3\,HCP, a Queen 2\,HCP and a Jack 1\,HCP. Information given by the players in the auction often relate to their number of HCP and their \emph{distribution} (the number of cards in one or more suits).

\end{itemize}

\subsection{Problem Statement} 

After receiving suggestions from bridge experts, we chose to analyse the following situation:
\begin{itemize}
\item West, the dealer, bids \Ss{4}, which (roughly) means that they have a minimum of 7 spades, and a maximum of 10\,HCP in their hand.
\item North Doubles, (roughly) meaning that they have a minimum of 13\,HCP and, unless they have a very strong hand, a minimum of three cards in each of the other suits (\Cs{}, \Ds{} and \Hs{}).
\item East passes, which has no particular meaning.
\end{itemize}
South must then make a decision: pass and let the opponents play $4\Ss{}^X$, or bid, and have their side play a contract. This is a high stakes decision that bridge experts are yet to agree on a precise formulation for. Our objective is to develop a methodology for representing this problem, and to find accurate and explainable solutions using relational learners. It should be noted that Derek Patterson was interested in solving this problem using genetic algorithms \cite{patterson}.

In the remainder of the article, we describe the various processes used in data generation, labelling, supervised learning, followed by a discussion of the results and the explicit models produced. These processes use relational representations of the objects and models involved, keeping bridge experts in the loop and allowing them to make adjustments where required.

In the next section we consider the relational formulation of the problem, and the data generation and labelling.

\section{Automatic Data Generation and Modelling Methodology}
\label{13genedonnees}
The methodology to generate and label the data consists of the following steps:
\begin{itemize}
\item Problem modelling
\item Automatic data generation
\item Automatic data labelling 
\item ILP framing
\end{itemize}
These steps are the precursors to running relational rule induction (Aleph) and decision tree induction (Tilde) on the problem.
\subsection{Problem Modelling} \label{13problemModelling}
The problem modelling begins by asking experts to define, in the context described above, two rule based models: one to characterize the hands such that West makes the \Ss{4} bid, and another to characterize the hands such that North makes the Double bid.  These rule based models are submitted to simulations allowing the experts to interactively validate their models. With the final specifications, we are able to generate examples for the target problem. For this section we introduce the terms:
\begin{itemize}
    \item \emph{nmpq exact distribution} which indicates that the hand has $n$ cards in \Ss{}, $m$ cards in \Hs{}, $p$ cards in \Ds{} and $q$ cards in \Cs{}, where $n+m+p+q=13$.
    \item \emph{nmpq distribution} refers to an exact distribution sorted in decreasing order (thus ignoring the suit information). For instance, a $2533$ exact distribution is associated to a $5332$ distribution.
    \item $|c|$ is the number of cards held in the suit $c\in\{\Ss{},\Hs{},\Ds{},\Cs{}\}$.
\end{itemize}

\subsubsection{Modelling the 4\Ss{} bid} \label{13fsBid}
Experts have modeled the 4\Ss{} bid by defining a disjunction of 17 rules that relate to West hand:
\begin{equation} \label{13fsRules}
4\Ss{} \leftarrow \bigvee\limits_{i=1}^{17} R_i, \mbox{ where } R_i = C_0 \wedge V_i \wedge C_i 
  \end{equation}
  in which 
\begin{itemize}
    \item $C_0$ is a condition common to the 17 rules and is reported in Listing \ref{13common-4s-rules} in the Appendix.
    \item $V_i$ is one of the four possible vulnerability configurations (no side vulnerable, both sides vulnerable, exactly one of the two sides vulnerable).
    \item $C_i$ is a condition specific to the $R_i$ rule.
\end{itemize}
For instance, the conditions for rules $R_2$ and $R_5$ are:
\begin{itemize}
	\item $V_2=$ East-West not vulnerable, North-South vulnerable.
	\item $C_2=$ all of: 
	\begin{itemize}
        \item $|\Hs{}|=1$,
       	\item  2 cards exactly among Ace, King, Queen and Jack of \Ss{}, 
        \item a 7321 distribution.
    \end{itemize}
	\item $V_5=$ East-West not vulnerable, North-South vulnerable.
	\item $C_5=$ all of: 
	\begin{itemize}
		\item $|\Cs{}| \geq 4$ or $|\Ds{}| \geq 4$, 
	    \item 2 cards exactly among Ace, King, Queen and Jack of \Ss{},  
	    \item a $7mpq$ distribution with $m\geq4$.
    \end{itemize}
\end{itemize}
To generate boards that satisfy these rules, we randomly generated complete boards (all 4 hands), and kept the boards where the West hand satisfies one of the 17 rules.
The experts were able to iteratively adjust the rules as they analysed boards that either passed through the filter, or failed to pass through the filter (but perhaps should have).
After the experts were happy with the samples, 8,200,000 boards were randomly generated to analyse rule adherence, and 10,105 of them contained a West hand satisfying one of the 17 rules. All rules were satisfied at least once. Of the times where at least one rule was satisfied, $R_2$, for example, was satisfied 16.2\% of the time, and $R_5$ was satisfied 15\% of the time. 

\subsubsection{Double Modelling}
Likewise, the bridge experts also modeled the North Double by defining a disjunction of 3 rules relating to the North hand:
\begin{equation} \label{13doubleRules}
\mbox{Double} \leftarrow \bigvee\limits_{i=1}^{3} R'_i, \mbox{ where }R'_i = C'_0 \wedge C'_i
\end{equation} 
This time, the conditions do not depend on the vulnerability.  The common condition $C'_0$ and the specific conditions $C'_i$ are as follows:
 \begin{itemize}
    \item $C'_0$ - for all $c, c_1, c_2\in\{\Hs{},\Ds{},\Cs{}\}$, $|c| \leq 5$ and not ($|c_1| = 5$ and $|c_2| = 5$).
    \item $C'_1$ - HCP $\geq$ 13 and $|\Ss{}| \leq$ 1.
    \item $C'_2$ - HCP $\geq$ 16 and $|\Ss{}| = 2$ and $|\Hs{}| \geq$ 3 and $|\Ds{}| \geq$ 3 and $|\Cs{}| \geq$ 3.
    \item $C'_3$ - HCP $\geq$ 20.
\end{itemize}
The same expert validation process was carried out as in Section \ref{13fsBid}. A generation of 70,000,000 boards resulted in 10,007 boards being satisfied by at least one 4\Ss{} bid rule for the West hand and at least one Double rule for the North hand. All rules relating to Double were satisfied at least once. Of the times that at least one rule was satisfied, $R'_1$, for example, was satisfied 69.3\% of the time, and $R'_2$ was satisfied 24.8\% of the time.  

\subsection{Data Generation}\label{13datageneration}
The first step of the data generation process is to generate a number of South hands in the context described by the 4\Ss{} and Double rules mentioned above. Note, again, that East's Pass is not governed by any rules, which is close to the real situation.

We first generated 1,000 boards whose West hands satisfied at least one \Ss{4} bid rule and whose North hands satisfied at least one  Double rule. One such board is displayed in Example \ref{13example1}:
\begin{ex}
\label{13example1}
 A board (North-South vulnerable / East-West not vulnerable)  which contains a  West hand satisfying rules $R_0$ and $R_5$ and a North hand satisfying rules $R'_0$ and $R'_1$:
{\footnotesize 
\begin{center}
\begin{bridgetable}

\begin{est}
\pique{10&3}
\coeur{J&8&4}
\carreau{A&7}
\trefle{Q&J&8&7&6&2}
\end{est}

\begin{ouest}
\pique{A&K&J&8&7&5&2}
\coeur{7}
\carreau{J&8&6&3}
\trefle{5}
\end{ouest}

\begin{sud}
\pique{Q&9&4}
\coeur{K&10&6&5&2}
\carreau{9&5}
\trefle{K&10&3}
\end{sud}

\begin{nord}
\pique{6}
\coeur{A&Q&9&3}
\carreau{K&Q&10&4&2}
\trefle{A&9&4}
\end{nord}

\end{bridgetable}
\end{center}
}
\end{ex}

For reasons that become apparent in Section \ref{13dataLabel}, for each of the 1000 generated boards, we generated an additional 999 boards. We did this by fixing the South hand in each board, and randomly redistributing the cards of the other players until we found a board satisfying Equations \ref{13fsRules} and \ref{13doubleRules}.
As a result of this process, 1,000  files were  created, with each file containing 1,000 boards that have the same South hand, but different West, North and East hands. 
\begin{ex}
\label{13example2}
One of the 999 other boards generated:
\begin{itemize}
\item West hand: \Ss{}AK108653 \Hs{}4 \Ds{}832 \Cs{}84 
\item North hand: \Ss{}J2 \Hs{}AJ73 \Ds{}AK106 \Cs{}AJ9 
\item East hand: \Ss{}7 \Hs{}Q98 \Ds{}QJ74 \Cs{}Q7652 
\item  South hand: \Ss{}Q94 \Hs{}K10652 \Ds{}95 \Cs{}K103. 
\end{itemize}
Note that the South hand  is identical to the one in Example \ref{13example1}, the West hand satisfies rules $R_0$ and $R_2$, and the North hand satisfies rules $R'_0$ and $R'_2$.
\end{ex}

\subsection{Data Labelling} \label{13dataLabel}

The goal is to label each of the 1,000 South hands associated to the 1,000 sample files, with one of following labels:
\begin{itemize}
\item \textbf{Pass} when the best decision of South is to pass (and therefore have West play 4\Ss{}$^X$).
\item \textbf{Bid} when the best decision of South is to bid (and therefore play a contract on their side).
\item \textbf{?} when the best decision is not possible to be determined. We exclude these examples from ILP experiments.
\end{itemize}

These labels were assigned by computing the score at 4\Ss{}$^X$ and other possible contracts by simulation.

\subsubsection{Scores Computation} 
During the playing phase, each player sees their own hand, and that of the dummy (which is laid face up on the table). A \emph{Double Dummy Solver} (DDS) is a computation (and a software) that calculates, in a deterministic way, how many tricks would be won by the declarer, under the pretext that everyone can see each other's cards. Though unrealistic, it is nevertheless a good estimator of the real distribution of tricks amongst the two sides \cite{pavlicek8J45}. For any particular board, a DDS can be run for all possible contracts, so one can thus deduce which contract will likely yield the highest score.

For each example file (which each contain 1,000 boards with the same South hand), a DDS software \cite{ddsrepo} was used to determine the score for the following situations:
\begin{itemize}
    \item 4\Ss{}$^X$ played by the East/West side.
    \item All available contracts played by the North/South side, with the exclusion of 4NT, 5NT, 5\Ss{}, 6\Ss{} and 7\Ss{} (experts deemed these contracts infeasible to be reached in practice). 
\end{itemize}

\subsubsection{Labels Allocation} The labels are assigned to each unique South hand by applying the following tests sequentially.
\begin{enumerate}
    \item Take the DDS score from the best North/South contract on each of the 1,000 boards (the optimal contract may not always be the same), and average them. If the average score of defending 4\Ss{}$^X$ is higher than this score, assign the label of \emph{Pass} to the South hand.
    \item Take the North/South contract with the highest mean DDS score over all of the boards (this may not be the best contract on each board, but merely the contract with the highest mean score across all boards). If the mean score of this contract is greater than the mean score at 4\Ss{}$^X$, assign \emph{Bid} to the South hand. If the mean score of the contract is at least $30$ total points lower than the mean score at 4\Ss{}$^X$, assign \emph{Pass}. Otherwise, assign the label \emph{?}.
\end{enumerate}

The initial dataset containing 1,000 sample files is reduced to a set $S$ of 961 examples after elimination of the 39 deals with the \emph{?} label. The resulting dataset consists of:
\begin{itemize}
    \item 338 \emph{Bid} labels, and
    \item 623 \emph{Pass} labels.
\end{itemize}

\subsection{Relational data modelling}
\label{13modrel}
Now that we have generated and labeled our data, the next step is to create the relational representations to be used in relational learners. The learning task consists in using what is  known about a given board in the context described above, and the associated South hand, and predicting what the best bid is. The best bid is known for the labeled examples generated in the previous section, but, of course, unknown for new boards.

We first introduce  the notations used in the remainder of the paper. 
\subsubsection{Logic Programming Notations}
\label{13lp-notations}
Examples are represented by terms and relations between terms.  Terms are \textit{constants}  (for instance the integers between 1 and 13), or atoms whose identifiers start with a lower-case character (for instance \textit{west}, \textit{east}, \textit{spade}, \textit{heart}, \ldots) or \textit{variables}, denoted by an upper-case character ($X$, $Y$, $Suit$, \ldots). Variables may instantiate to any constant in the domain. Relations between objects are described using \textit{predicates} applied to terms. A \textit{literal} is a predicate symbol applied to terms, and a \textit{fact} is a  ground literal,  i.e. a literal without variables, whose arguments are ground terms only. For instance, \textit{decision([sq, s9, s4, hk, h10, h6, h5, h2, d9, d5, ck, c10, c3], 4, n, west, bid)} is a ground literal of predicate symbol \textit{decision}, with four arguments, namely, a hand described as a list of cards,  and 4 constant arguments (see Section \ref{13targetPredicate} for further details).

In order to introduce some flexibility in the problem description, the \textit{background knowledge} (also known as the \textit{domain theory})  is described as set of \textit{definite clauses}. We can think of a definite clause as a first order logic rule, containing a head (the conclusion of the rule) and a body (the premises of the rule). For instance, the following clause: \\
$nb(Hand,Suit,Value) \leftarrow suit(Suit), count\_cards\_of\_suit(Hand,\allowbreak Suit, \allowbreak Value)$\\
has the head $nb(Hand,Suit,Value)$ where $Hand$, $Suit$ and $Value$ are variables  and two literals in the body  $suit(Suit)$  and  $count\_cards\_of\_suit(Hand,\allowbreak Suit,\allowbreak Value)$.
The declarative interpretation of the rule is that given that variable $Suit$ is a valid bridge suit, and given that $Value$ is the number of cards of the suit $Suit$ in the hand $Hand$, one can derive, in accordance with the logical consequence relationship $\models$ \cite{LLOYD-ILP} that $nb(Hand,Suit,Value)$ is true. 
This background knowledge will be used to derive additional information concerning the examples, such as hand and board properties. 

Let us now describe the \emph{decision} predicate, the target predicate of the relational learning task.

\subsubsection{Target Predicate}\label{13targetPredicate} 
Given a hand, the position of the hand, vulnerability, and the dealer's position, the goal is to predict the class label. The target predicate for the bidding phase  is $\mathit{decision}(Hand,\allowbreak \mathit{Position},\allowbreak \mathit{Vul},\allowbreak \mathit{Dealer},\allowbreak \mathit{Class})$ where the arguments are variables. 
\begin{itemize}
    \item $\mathit{Hand}$: the 13 cards of the player who must decide to \emph{bid} or \emph{pass}.
    \item $\mathit{Position}$: the relative position of the player to the dealer. In this task, \emph{Position} is always 4, the South hand.
    \item $\mathit{Vul}$: the vulnerability configuration ($b$ = both sides,  $o$ = neither side, $n$ = North-South or $e$ = East-West).
    \item $\mathit{Dealer}$: the cardinal position of the dealer  among \emph{north}, \emph{east}, \emph{south} or \emph{west}. In this task, \emph{Dealer} is always \emph{west}.
    \item $\mathit{Class}$: label to predict (\emph{bid} or pass).
\end{itemize}
In this representation, a labeled example is a grounded positive literal containing the decision predicate symbol.
Example 1 given in section \ref{13datageneration} is therefore represented as: \\
$decision([sq,s9,s4,hk,h10,h6,h5,h2,d9,d5,ck,c10,c3],4,n,west,bid).$
Such a ground literal contains all the explicit information about the example. However, in order to build an effective relational learning model of the target predicate as a set of rules, we need to introduce abstractions of this information, in the form of instances of additional predicates in the domain theory. These predicates are described below.

\subsubsection{Target concept predicates}

In order to learn a general and explainable model for the target concept, we need to carefully define the target concept representation language by choosing  predicate symbols that can appear in the model, more specifically, in the body of definite clauses for the target concept definition. We thus define new predicates in the domain theory, both extensional (defined by a set of ground facts) and intensional (defined by a set of rules or definite clauses). The updated domain theory allows to complete the example description, by gathering/deriving true facts related to the example. It also implicitly defines a target concept language that forms the search space for the target model which the Inductive Logic Programming algorithms will explore using different strategies. Part of the domain theory for this problem was previously used in another relational learning task for a bridge bidding problem \cite{ilpbridgegamechallenge}, illustrating the reusability of  domain theories.  The predicates in the target concept language are divided in two subsets:

\begin{itemize}
\item Predicates that were introduced in \cite{ilpbridgegamechallenge}. These include both general predicates, such as $gteq(A, B)$ ($A\geq B$), $lteq(A, B)$ ($A \leq B$) and predicates specific to bridge, such as $nb(Hand, Suit, Number)$, which tests that a suit $Suit$ of a hand $Hand$ has length $Number$ (e.g. $nb(Hand,heart,3)$ means that the hand $Hand$ has 3 cards in \Hs{})

\item Higher level predicates not used in \cite{ilpbridgegamechallenge}. Among which:
	\begin{itemize}
	\item $hcp(Hand,Number)$ is the number of High Card Points (HCP) in $Hand$.
	\item $suit\_representation(Hand,Suit,Honors,Number)$ is an  abstract representation of a suit.  $Honors$ is the list of cards strictly superior to 10 of the suit $Suit$ in hand $Hand$. $Number$  is the total number of cards in $Suit$ suit.
	\item $distribution(Hand, [N,M,P,Q])$ states that the $nmpq$ distribution (see Section \ref{13problemModelling}) of hand $Hand$ is the ordered list $[N,M,P,Q]$. 

	\end{itemize}
\end{itemize}

For instance, in Example 1 from Section \ref{13example1}, we  may add to the representation the following literals:
 \begin{itemize}
    \item $hcp([sq,s9,s4,hk,h10,h6,h5,h2,d9,d5,ck,c10,c3],8).$
    \item $suit\_representation([sq,s9,s4,hk,h10,h6,h5,h2,d9,d5,ck,c10,c3],spade,[q],3).$
    \item  $distribution([sq,s9,s4,hk,h10,h6,h5,h2,d9,d5,ck,c10,c3], [2,3,3,5]).$
\end{itemize}

Note again, that a fact (i.e. ground literal) can be independent of the rest of the domain theory, such as $has\_suit(hk,heart)$ which states that $hk$ is a  \Hs{} card, or it can be inferred from existing facts and clauses of the domain theory, such as
$nb([sq,s9,s4,hk,h10,h6,h5,h2,d9,d5,ck,c10,c3],heart,5)$, which derives from:
\begin{itemize}
\item $suit(heart)$:  $heart$ is a valid bridge suit.
\item $has\_suit(hk,heart)$: hk is a $heart$ card.
\item $nb(Handlist,Suit,Value) \leftarrow suit(Suit), count\_card\_of\_suit(Handlist,\allowbreak Suit,\allowbreak Value)$ where $count\_card\_of\_suit$  has a recursive definition   within the domain theory.
\end{itemize}

\section{Learning Expert Rules}
\label{13ilp}
\subsection{ Inductive Logic Programming Systems}
The ILP systems used in our experiments were \emph{Aleph} and \emph{Tilde}. These are two mature state of the art 
ILP systems of the \emph{symbolic  SRL}  paradigm\footnotemark. 
Recent work showed that such ILP systems,  Tilde in particular, are competitive with recent \emph{distributional SRL}  
approaches that first learn a \emph{knowledge graph embedding} to encode the relational datasets into score matrices  
that can then fed to non-relational learners. Given a range of some relational datasets, \cite{Dumancic:2019vf} 
shows that there is no clear winner on  classification tasks when comparing distributional and symbolic learning 
methods,  and that both exhibit  strengths and limitations. 
Tilde on the other hand outperforms 
KGE based approaches on Knowledge Base Completion tasks (namely concerning the ability to infer the status (true/false) for missing facts in relational datasets). 

As our work aims to build explainable models in a deterministic context we only consider here  ILP systems.
The most prominent difference between Aleph and Tilde is   their output:  a set of independent relational rules for Aleph and a relational decision tree for Tilde. One important declarative parameter 
of both systems   -- and in symbolic  SRL systems in general --  is the so-called \emph{language bias} that specifies 
which predicates defined in the background knowledge may be used
 and in what form these predicates  appear in the learned model (the rule set for Aleph and the decision tree for Tilde)
\footnotetext{SRL stands for \emph{Statistical Relational Learning} and is a recent development of ILP \cite{DeRaedt2016} that integrates learning and reasoning under uncertainty about individuals and actions effect.} 
\subsubsection{Aleph}
Given a set of positive examples $E^+$, a set of negative examples $E^-$, each as ground literals of the target
concept predicate and a domain theory $T$, Aleph \cite{aleph} builds a \emph{hypothesis} $H$  as a logic program, i.e. a set of definite clauses, such that $\forall e^+ \in E^+: H \wedge T \models e^+$ and $\forall e^- \in E^- : H \wedge T \not \models e^-$. In other words, given the domain theory $T$, and after learning hypothesis $H$, it is possible  from each positive example description $e^+$ to derive that $e^+$ is indeed a positive example, while such a derivation is impossible for any of the negative examples $e^-$.  
 The domain theory $T$ is represented by a set of facts, i.e. ground positive literals from other predicates and clauses, possibly inferred from a subset of primary facts and a set of   unground clauses (see Section \ref{13modrel}).
$H$ is a set of definite clauses where the head predicate of each rule is the target predicate.  An example of  such a hypothesis $H$ is given in Listing \ref{13complete-best-rules-aleph}.

\subsubsection{Tilde}
Tilde\footnotemark \cite{Blockeel:1998aa} relies on the
\emph{learning by interpretation} setting.
\footnotetext{Available as part of the ACE Ilp system at \url{https://dtai.cs.kuleuven.be/ACE/}}
Tilde learns  a relational binary decision tree in which the nodes  are conjunctions of literals that can share variables  with the following restriction: a variable introduced in a node cannot appear in the right branch below that node (i.e. the failure branch of the test associated to the node). In the decision tree, each example, described through a set of ground literals,  is propagated through the current tree until it reaches a leaf. In the final tree, each leaf is associated to the decision \emph{bid}/\emph{pass}, or a probability distribution on these decisions.  An example of a Tilde decision tree output is given in Listing \ref{13best-tree-tilde-complete}.


\subsection {Learning setup}
\label {13expes}
In our problem, the target predicate is \emph{decision}, as defined in Section \ref{13targetPredicate} and exemplified in Section \ref{13targetPredicate}. Regarding  Aleph, in  positive examples the last argument of the target predicate has value \emph{bid} while in negative examples it has value \emph{pass}. 
The  Aleph standard strategy \emph{induce} learns the logic program as a set of clauses  using a hedging strategy: at each iteration it selects a  seed positive example $e$ not entailed  by the current solution,  
find a clause $R_e$ which covers $e$, and then remove  from the learning set the positive examples covered by $R_e$. The learning procedure is repeated until all positive examples are covered. The \emph{induce} strategy is sensitive to the order of the learning examples.
We also used Aleph with \emph{induce\_max}, which is more expensive but insensitive to the order of positive examples: it constructs a best clause, in terms of coverage,  for each positive example, then selects a clause subset. \emph{induce\_max} has proved beneficial in some of our experiments. 

\subsubsection{Experimental  setup}
The performance of the learning system with respect to  the size of the training set is averaged over $50$ executions $ i $. The sets $ Test_i $ and $ Train_i $ are built  as follows:
\begin {itemize}
	\item $Test_i = 140$ examples randomly selected from $S$ using seed $i$.
	\item Each $ Test_i $ is stratified, i.e. it has the same ratio of examples labeled \emph{bid}  as  $S$  (35.2 \%).
	\item $ Train_i = S \setminus Test_i $ (820 examples).
\end {itemize}

For each execution $ i $, we randomly generate from $Train_i$ a sequence of increasing number of examples $ n_k = 10 + 100 \times k $ with k between 0 and 8. We denote each of these subsets $ T_ {i, k} $, so that $|T_{i,k}|=n_k$ and $ T_ {i, k} \subset T_ {i, k + 1}$.
Each model trained on $ T_ {i, k} $ is evaluated on $ Test_i $.  

\subsubsection{Evaluation criteria}
As we can interpret our methodology as a post-hoc agnostic black-box explanation framework, we evaluate this 
frame\-work in terms of \emph{fidelity}, \emph{unambiguity} and \emph{interpretability}\cite{LakkarajuKCL19}. \begin{itemize}
\item 
A  high \emph{fidelity} explanation should faithfully mimic 
the behavior of the black box model. 
To measure fidelity  we  use  the  \emph{accuracy} of the surrogate relational model
with respect to the he labels assigned by the black box model.
\item \emph{Unambiguity} evaluates if one or multiple (potentially conflicting) explanations are provided by the relational surrogate model for a given instance. While we do not explicitly report on unambiguity in our experiments, we note that decision trees by nature are non ambiguous models. Rule sets, on the other hand,  may indeed display some overlap. However, but as  our Aleph experiments only consider  one target label (see Listing \ref{13complete-best-rules-aleph}) even if two rules apply for some instance, they conclude on the same label and the  model again displays no ambiguity.  That does not mean that the notion is useless in our context, rather that we have to find more sophisticate ways to evaluate in which sense such models include some level of ambiguity. 
\item  \emph{Interpretability}  quantifies how easy it is to understand and reason about the explanation. This dimension 
of explanation is  subjective and as such, maybe prone to the expert/user bias. We therefore choose a
crude but objective metric: the \emph{model complexity}, i.e. the number of rules/nodes of the models. 
We will also see that  in our experiments we obtain similar decision tree accuracies with $L_2$, a language containing only predicates appearing in a human-made model, as with the much larger language  $L_1$. Clearly, the $L_2$ models have better interpretability.
\end{itemize}

\subsection{First experiments}
\begin {figure}
\begin {center}
\begin {tabular} {cc}
\includegraphics [width = 0.5 \textwidth] {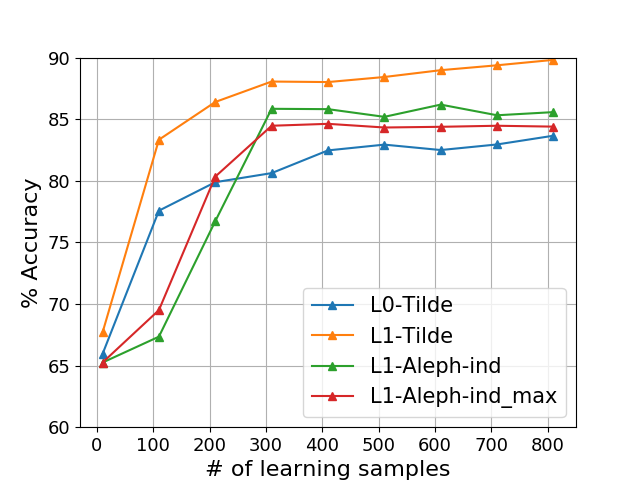}
&
\includegraphics [width = 0.5 \textwidth] {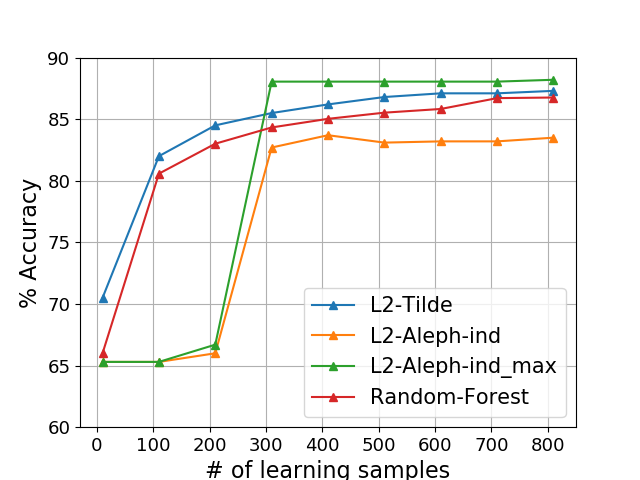}
\end {tabular}
\end {center}
\caption {Average model accuracies of $L_0$, $L_1$ models (left) and $L_2$ models (right).}
\label {13accuracy_chart}
\end {figure}

\begin {figure}
\begin {center}
\begin {tabular} {cc}
\includegraphics [width = 0.5 \textwidth] {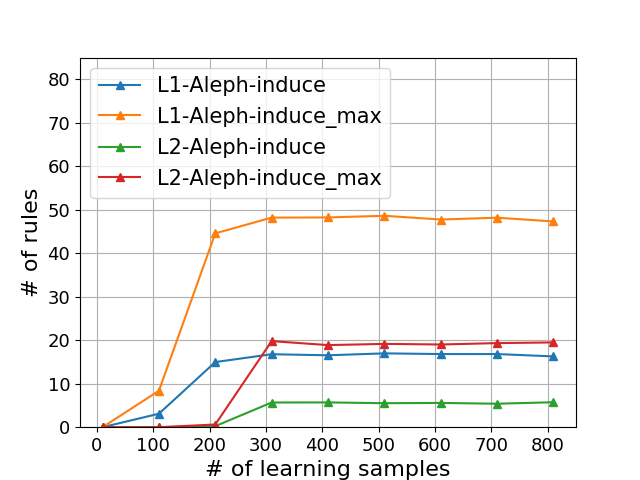}
&
\includegraphics [width = 0.5 \textwidth] {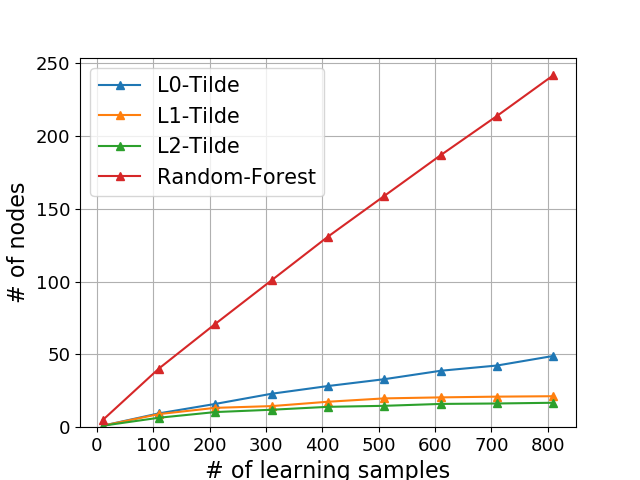}
\end {tabular}
\end {center}
\caption {Average model complexities of rule-based (left) and tree-based (right) models.}
\label {13complexity_chart}
\end {figure}

The first  experiments involve  a baseline non relational data representation,  together with the $L_0$ language mentioned earlier as well as a smaller languages $L_1$.  Table \ref{13summary_table}  and Figures  \ref{13accuracy_chart} and \ref{13complexity_chart}   display the results of our experiments  as well as  results obtained using a different language $L_2$ further introduced in Section \ref{newExperiments}.

We have first   learned  Random Forest models, using the \emph{sklearn} python package\cite{Pedregosa:2011tb},  with   default parameters including using forests made of 100 trees. The  data representation is  as follows: each hand is described by 54 attributes: vulnerability, target class, highest card, second highest card, \dots, $13^{\mathit{th}}$ highest card  in each suit (ordered as $\Cs{} < \Ds{} < \Hs{} < \Ss{}$). 
 If a player has  $n$ cards in a given suit, the corresponding $n+1, \dots ,13^{th}$ attributes for this suit are set to 0.

The relational  models are learned with Tilde and  Aleph where Aleph is trained on both the  \emph{induce} and \emph{induce\_max} settings.  Figure \ref{13accuracy_chart}-left presents the average accuracy, over all $ i = 1 \dots 50 $ executions, of the models trained on the sets $ T_ {i, k} $ and therefore requiring  building 450 models for each curve. Figure  \ref{13complexity_chart} displays the corresponding model complexity, expressed in average number of nodes for Tilde and Random Forest\footnotemark.
  \footnotetext{For Random Forest, the reported number of nodes  is an average over the 100 trees of a model}
  Table  \ref{13summary_table}
displays the average accuracies and complexities  of each experiment for learning set sizes 310 (k=4) and 810 (k=9). 
An experiment represents 450 Tilde or Aleph executions, and the resulting CPU cost of these executions is also recorded. As the experiments were performed on similar  machines with a different number of cores (either 8 or 16) we report  the CPU cost  as the  number of cores of the  machine times   the total  CPU time, i.e. in  \emph{core hours} units\footnotemark.
\footnotetext{For instance, 16 core-hours represents 1 hour of CPU time on a 16-core machine or 2 hours CPU time on a 8-core machine.}

\begin {table}[htb!]
\includegraphics [width = \textwidth] {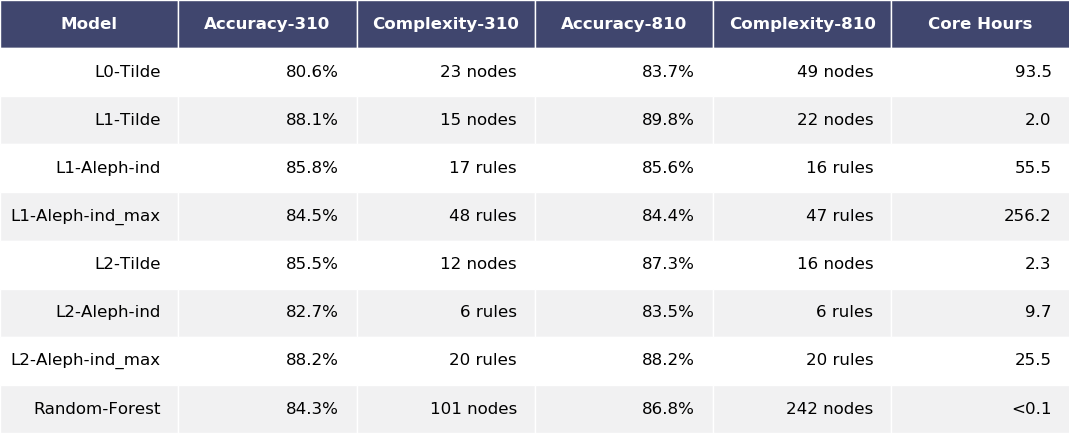}
\caption {Summary of average accuracy and complexity for all models when trained on 310 and 810 examples, together with total  CPU cost of the 450 Tilde or Aleph executions in $\mathit{CPU time} \times \mathit{hours}$. Reported complexity of Random Forest only concerns one tree among the 100 trees in the forest.
}
\label {13summary_table}
\end {table}

Aleph failed to cope with the large $L_0$ language, and couldn't return a solution in reasonable time\footnotemark. As can be seen in Figure \ref{13accuracy_chart} and Table \ref{13summary_table}, the $L_0$-Tilde models had rather low accuracy and a large number of nodes (which grew linearly with the number of training examples used).

\footnotetext{Aleph was able to learn models after we removed the $nb$ predicate from $L_0$, albeit rather slowly ($\approx384$ core hours).
}

After an analysis of the Tilde output trees, it became apparent that Tilde was selecting many nodes with very specific conditions, in particular, with the \emph{suit\_representation} and \emph{distribution} predicates. While this improved performance in training, the specificity hindered the performance in testing. To counter this, we removed from $L_0$ the \emph{suit\_representation} and \emph{distribution} predicates, creating the more concise language $L_1$.

As expected, the average number of nodes in $L_1$-Tilde reduced dramatically. $L_1$-Tilde had the best accuracy of all models, which continued to increase between $610$ and $810$ examples. Its complexity was relatively stable, increasing only slightly between $510$ and $810$ examples, to a total of $21$ tree nodes.
With the smaller $L_1$ language, $L_1$-Aleph-induce was able to find solutions in reasonable time. It's accuracy was worse than that of $L_1$-Tilde ($85.6\%$ compared to $89.8\%$ when training on $810$ examples). After $310$ examples, the $L_1$-Aleph-induce model size remained stable at $16$ rules (its accuracy also showed stability over the same training set sizes).

We also used  Aleph's \emph{induce\_max} strategy with $L_1$. Compared to \emph{induce}, this search strategy has a higher time cost and can produce rules with large overlaps in  coverage of learning examples, but can generalize better due to the fact that it saturates every example in the training set, not just a subset based on the coverage of a single seed example. This search strategy was of no benefit in this case, with $L_1$-Aleph-induce\_max showing slightly worse accuracy ($84.4\%$ compared to $85.6\%$) and a much higher complexity ($47$ rules compared to $15$) when compared with $L_1$-Aleph-induce.

The CPU cost of learning  the Random Forest models is negligible  (less than 10 seconds) compared to those of relational models. The accuracies lie between those of $L_1$-Tilde and of $L_1$-Aleph models, while the number of nodes per tree is much higher (see Figure \ref{13complexity_chart}-right). This indicates that there is no gain in using ensemble learning on non-relational decision trees, although the data representation is much simpler and needs only limited Bridge expertise to define.
The Random Forest models are, as expected,  difficult to interpret, apart from variable importance:  It appears that the first three highest cards in each suit and 
vulnerability are the most important variables in this task. It would be possible to use post-hoc  formal methods to \emph{a posteriori} explain those random forest models \cite{Izza2021},  however the whole process would be expensive, and, after \cite{Rudin2021}, \textit {Explaining black boxes, rather than 
replacing them with interpretable models, can make the problem worse by providing misleading or false characterizations}. 


Following these experiments, we had bridge experts do a thorough analysis of one of the successful $L_1$-Tilde outputs. Their goal was to create a sub-model which was much easier to interpret than the large decision trees\footnotemark, and had an accuracy comparable to that seen by $L_1$-Tilde. \footnotetext{The experts noted that, in general, the Tilde trees were more difficult to read than the rule-based output produced by Aleph.} We also used the experts' insights to further revise the domain theory, in the hope of learning a simpler model with Aleph and Tilde. 
We describe this process in the next section.


\section {Expert Rule Analysis and new Experiments}
\label {13analyse}
\subsection{Expert Rule Analysis}
To find an $L_1$-Tilde output tree for the experts to analyse, we chose a model with good accuracy and low complexity (a large tree would be too hard to parse for a bridge expert). A subtree of the chosen $L_1$-Tilde output is displayed in Listing \ref{13best-tree-tilde} (the full tree is displayed in Listing \ref{13best-tree-tilde-complete} in the Appendix). Despite the low complexity of this tree (only 15 nodes), it remains somewhat difficult to interpret.

\noindent
\begin{minipage}{\linewidth}
{\footnotesize
 \begin{lstlisting}[frame = single, caption = Part of the L1-tree learned by Tilde, label = 13best-tree-tilde]
decision(-A,-B,-C,-D,-E)
nb(A,-F,-G),gteq(G,6) ? 
+--yes: hcp(A,4) ? 
|       +--yes: [pass]
|       +--no:  [bid]
+--no:  hcp(A,-H),gteq(H,16) ? 
        +--yes: [bid]
        +--no:  nb(A,spade,-I),gteq(I,1) ? 
                +--yes: lteq(I,1) ? 
                |       +--yes: hcp(A,-J),lteq(J,5) ? 
                |       |       +--yes: [pass]
                |       |       +--no:  nb(A,-K,5) ? 
                |       |               +--yes: hcp(A,6) ? 
                |       |               |       +--yes: [pass]
                |       |               |       +--no:  [bid]
                |       |               +--no:  [pass]
\end{lstlisting}
}
\end{minipage}

Simply translating the tree to a decision list does not solve the issue of interpretability. Thus, with the help of bridge experts, we proceeded to do a manual extraction of rules from the given tree, following these post-processing steps:
\begin {enumerate}
\item Natural language translation from relational representations.
\item 
Removal of nodes with weak support (e.g. a node which tests for precise values of HCP instead of a range).
\item Selection of rules concluding on a \emph{bid} leaf.
\item Reformulation of intervals (e.g. combining overlapping ranges of HCP or number of cards).
\item Highlighting the specific number of cards in \Ss {}, and specific $nmpq$ distributions.
\end {enumerate}

The set of five rules resulting from this  post-processing, shown   Listing \ref{13best-rules} in the next section, are much more readable than $L_1$-Tilde trees. Despite its low complexity, this set of rules has an accuracy of $90.6\%$ on the whole of $S$.

\subsection{New experiments}\label{newExperiments}
Following this expert review, we created the new $L_2$ language, which only used the predicates that appear in the post-processed rules above. This meant the reintroduction of the \emph{distribution} predicate, the addition of the \emph{nbs} predicate (which is similar to \emph{nb} but denotes only the number of \Ss{} in the hand), and the omission of several other predicates. We display Table~\ref{13domain_theories} in the Appendix  the predicates involved  in the  languages $L_0$, $L_1$ and $L_2$.
Results are displayed together withe those regarding $L_0$ and $L_1$  on  Figure \ref{13accuracy_chart}, Figure \ref{13complexity_chart} and Table \ref{13summary_table}.

Using $L_2$ with Aleph-induce results in a much simpler model than $L_1$-Aleph-induce ($6$ rules compared to $16$ when trained on 810 examples), at the cost of somewhat lower accuracy ($83.5\%$ compared to $85.6\%$). The complexity of $L_2$-Tilde was also smaller than its counterpart $L_1$-Tilde (12 nodes compared to 15), also at the expense of accuracy ($87.3\%$ compared to $89.8\%$). In both cases, the abstracted and simplified $L_2$ language produced models which were far less complex (and thus more easily interpretable), but had a slight degradation in accuracy, when compared with $L_1$.

When using $L_2$ the more expensive induce\_max search strategy proved useful, with $L_2$-Aleph-induce\_max showing an accuracy of $88.2\%$, second only to $L_1$-Tilde ($89.8\%$). This was at the cost of complexity, with $L_2$-Aleph-induce\_max models having an average of 20 rules compared to 6 for $L_2$-Aleph-induce.

We selected a $L_2$-Aleph-induce\_max model, displayed   Listing \ref{13complete-best-rules-aleph} in the Appendix. The rules of this model are relatively easy to apprehend, and the model achieves an accuracy of $89.7\%$ on the whole example set $S$.  We provide Listing \ref{13best-rules-aleph-jp} the experts translation $M$  of this model and compare it to  the rewriting $H$ displayed Listing \ref{13best-rules}  of the $L_1$-Tilde tree model which led to the $L_2$ language definition. The translation is much simpler that the previous one and required only steps 1,4 and 5 of the original translation process. An in-depth expert explanation of how the translated models  differ is provided in the Appendix. It is noticeable that the two translated models are very similar:  each rule in $M$ matches, with only slight differences, the rule in $H$ with same number. The only apparent exception is the overly specific  rule $M2b$. A closer look to rule $H2$ reveals that rule $M2b$ completes rule $M2$ by considering the case in which the player has $6$ cards, leading to a closer match to human rule $H2$.    
%
%
%

\noindent
\begin{minipage}{\linewidth}
{\footnotesize 
\begin{lstlisting}[frame = single, caption = Translation  of the $L_2$-Aleph-induce\_max model, label = 13best-rules-aleph-jp, escapechar = \%]
We bid: 
M1  - With 0%\Ss{}.%. 
M2  - With 7+ cards in a non-spade suit OR 14+HCP.
M2b - With 6+ cards in a non-spade suit AND 
	(1%\Ss{}% AND 5-13HCP)   OR
	(2%\Ss{}% AND 7-13HCP)   OR
	(3%\Ss{}% AND 10-13HCP)  
M3  - With 1%\Ss{}%  AND (5-4-3-1%\Ss{}% OR 5-5-2-1%\Ss{}%) AND 8-13HCP.
M4  - With 2%\Ss{}%  AND  5-5-2%\Ss{}%-1
M5  - With 2%\Ss{}%  AND (5-4-2-2%\Ss{}% OR 5-3-3-2%\Ss{}%) AND 13HCP.
\end{lstlisting}
}
\end{minipage}

\noindent
\begin{minipage}{\linewidth}
{\footnotesize 
\begin{lstlisting}[frame = single, caption = Rules extracted by the experts from the L1-Tilde tree, label = 13best-rules, escapechar = \%]
We bid: 
H1. - With 0%\Ss{}%.
H2  - With 6+ cards in a non-spade suit   OR 16+HCP.
H3  - With 1%\Ss{}% AND  (5-4-3-1%\Ss{}% or 5-5-2-1%\Ss{}%) AND 6-15HCP.
H4  - With 2%\Ss{}% AND   5-5-2%\Ss{}%-1              AND 7-15HCP.
H5  - With 2%\Ss{}% AND   5-4-2-2%\Ss{}%              AND 11-15HCP.
\end{lstlisting}
}
\end{minipage}

\section{Conclusion}
\label{13conclusion}
We conducted  an end-to-end experiment to acquire expert rules for a bridge bidding problem. 
We did not propose  new technical tools, our contribution was in investigating a complete methodology to handle a learning task from beginning to end, i.e.  from the specifications of the learning problem to a learned model,  obtained after expert feedback, who not only criticise and evaluate the models, but also propose hand-made (and computable)  alternative models. The methodology includes generating and labelling  examples by using both domain experts (to drive the modelling and relational representation of the examples) and a powerful simulation-based black box  for labelling. The result is a set of high quality examples, which is a pre\-re\-quisite for learning a compact and accurate  relational model. The final relational model is built after an additional phase of interactions with experts, who investigate  a learned model, and propose an alternative model, an analysis of which leads to further refinements of the learning language.  We argue that  interactions between experts and engineers are made easy when the learned models are relational, allowing experts to revisit and contradict  them. Indeed, a detailed comparison between  the models, as translated by experts, shows that the whole process has led to quite a fixed-point. 
Clearly, learning tasks in a well defined and simple universe with precise and deterministic  rules, such as in games,  are good candidates for the interactive methodology we proposed. The extent to which experts and learned models cooperating in this way can be transferred to less well defined problems, including, for instance, noisy environments, has yet to be investigated. Note that complementary techniques can also be applied, such as active learning, where machines or humans dynamically provide  new examples to the system, leading to revised models. Overall, our contribution is a step towards a complete handling of learning tasks for decision problems, giving domain experts (and non-domain experts alike) the tools to both make accurate decisions, and to understand and scrutinize the logic behind those decisions.

\bibliographystyle{apalike}
\bibliography{egc2020_bis}
%
\appendix
\section*{Appendix 1}
\section{Tree-based and rule-based models}
\noindent
\begin{minipage}{\linewidth}
{\footnotesize 
 \begin{lstlisting}[frame = single, caption = A complete L1-tree learned by Tilde, label = 13best-tree-tilde-complete]
decision(-A,-B,-C,-D,-E)
nb(A,-F,-G),gteq(G,6)? 
+--yes: hcp(A,4)? 
|   +--yes: [pass] 2.0
|   +--no:  [bid] 57.0
+--no:  hcp(A,-H),gteq(H,16)? 
    +--yes: [bid] 6.0
    +--no:  nb(A,spade,-I),gteq(I,1)? 
        +--yes: lteq(I,1)? 
        |   +--yes: hcp(A,-J),lteq(J,5)? 
        |   |   +--yes: [pass] 11.0
        |   |   +--no:  nb(A,-K,5)? 
        |   |       +--yes: hcp(A,6)? 
        |   |       |   +--yes: [pass] 3.0
        |   |       |   +--no:  [bid] 27.0
        |   |       +--no:  [pass] 3.0
        |   +--no:  hcp(A,15)? 
        |       +--yes: [bid] 3.0
        |       +--no:  nb(A,-L,3)? 
        |           +--yes: [pass] 146.0 
        |           +--no:  hcp(A,11)? 
        |               +--yes: [bid] 3.0 
        |               +--no:  hcp(A,-M),lteq(M,6)? 
        |                   +--yes: [pass] 15.0 
        |                   +--no:  lteq(I,2)? 
        |                       +--yes: nb(A,-N,1)? 
        |                       |   +--yes: [bid] 3.0
        |                       |   +--no:  hcp(A,-O),gteq(O,11)? 
        |                       |       +--yes: [bid] 5.0
        |                       |       +--no:  [pass] 16.0
        |                       +--no:  [pass] 5.0
        +--no:  [bid] 5.0 

\end{lstlisting}
}
\end{minipage}

\noindent
\begin{minipage}{\linewidth}
{\footnotesize 
\begin{lstlisting} [frame=single,caption=A $L_2$-Aleph-induce\_max model,  label=13complete-best-rules-aleph,escapechar=\%]
decision(A,4,B,C,bid) :-
   nbs(A,0).
decision(A,4,B,C,bid) :-
   hcp(A,D), gteq(D,5), nb(A,E,F), gteq(F,6), nbs(A,1).
decision(A,4,B,C,bid) :-
   hcp(A,D), gteq(D,9), nbs(A,1).
decision(A,4,B,C,bid) :-
   hcp(A,D), gteq(D,8), nb(A,E,F), gteq(F,5), nbs(A,1).
decision(A,4,B,C,bid) :-
   hcp(A,D), gteq(D,4), distribution(A,[6,4,2,1]).
decision(A,4,B,C,bid) :-
   hcp(A,D), gteq(D,8), nb(A,E,F), gteq(F,6), nbs(A,2).
decision(A,4,B,C,bid) :-
   hcp(A,D), gteq(D,13), nb(A,E,F), gteq(F,5), nbs(A,2).
decision(A,4,B,C,bid) :-
   hcp(A,D), gteq(D,10), nb(A,E,F), gteq(F,6).
decision(A,4,B,C,bid) :-
   nb(A,D,E), gteq(E,7).
decision(A,4,B,C,bid) :-
   hcp(A,D), gteq(D,14), nb(A,E,F), gteq(F,5).
decision(A,4,B,C,bid) :-
   nb(A,D,E), lteq(E,1), nbs(A,2).
decision(A,4,B,C,bid) :-
   hcp(A,D), nb(A,E,D), nb(A,F,G), gteq(G,6).
decision(A,4,B,C,bid) :-
   hcp(A,D), gteq(D,12), nb(A,E,F), lteq(F,1).
decision(A,4,B,C,bid) :-
   hcp(A,D), lteq(D,7), gteq(D,7), nb(A,E,F), gteq(F,6).
decision(A,4,B,C,bid) :-
   hcp(A,D), gteq(D,14), nbs(A,2).

\end{lstlisting}
}
\end{minipage}

\begin{table}[htb!]
  \begin{center}
    \begin{tabular}{c|c|c} 
      \pmb{$L_0$} & \pmb{$L_1$} & \pmb{$L_2$}\\
      \hline
      $suit\_rep(H,Suit,Hon,Num)$ & & \\
      $distribution(H,[N,M,P,Q])$ &  & $distribution(H,[N,M,P,Q])$\\
      $nb(H,Suit,Num)$ & $nb(H,Suit,Num)$ & $nb(H,Suit,Num)$\\
      $hcp(H,Num)$ & $hcp(H,Num)$ & $hcp(H,Num)$\\
      $semibalanced(H)$ & $semibalanced(H)$ & 
      \\
      $unbalanced(H)$ & $unbalanced(H)$ & 
      \\
      $balanced(H)$ &
      $balanced(H)$ & 
      \\
      $vuln(Pos,Vul,Dlr,NS,EW)$ & $vuln(Pos,Vul,Dlr,NS,EW)$ &
      \\
      $vuln(Pos,Vul,Dlr,NS,EW)$ & $vuln(Pos,Vul,Dlr,NS,EW)$ &
      \\
      $lteq(Num,Num)$ & $lteq(Num,Num)$ &
      $lteq(Num,Num)$\\
      $gteq(Num,Num)$ & $gteq(Num,Num)$ &
      $gteq(Num,Num)$\\
      $longest\_suit(H,Suit)$ &
      $longest\_suit(H,Suit)$ &
      \\
      $shortest\_suit(H,Suit)$ &
      $shortest\_suit(H,Suit)$ &
      \\
      $major(Suit)$ &
      $major(Suit)$ &
      \\
      $minor(Suit)$ &
      $minor(Suit)$ &
      \\
      & & $nbs(H,Num)$\\
    \end{tabular}
    \caption{Predicates in the domain theories.}\label{13domain_theories}
  \end{center}
\end{table}
\section*{Appendix 2}
\section{Expert comparison of models $M$ and $H$}
From an expert point of view, the  model $H$  from  the $L_1$-Tilde model (see Listing \ref{13best-rules}) and  the model $M$ from the $L_2$-Aleph-InduceMaxrules (see Listing \ref{13best-rules-aleph-jp}), both obtained after a human post-processing, are very similar. They differ only on transition zones of certain characteristics, which are of low support. A detailed analysis follows.

\subsection{Transition zones within HCP}  The models differ on the range 14-15 HCP (M2 vs H2). Note that with very unbalanced hand distributions, some parameters  will be of a greater influence than a deviation of 1 or 2HCP to take a decision: we can mention for example the number of HCP outside the \Ss{} suit or the number of HCP in the 4-card-suits (or more). An increase in the associated values would be an indication that, for the same value of HCP, the position of honors in the hand of the player is more favorable to the decision \emph{bid}.

\subsection{Transition zones within distribution} 
A   M2b rule  deals with hands holding a 6-card-suit exactly while those which have a 7-card-suit generate the same decision \emph{bid}.  Within $ L_1$ and $L_2$ models  hands with 5-card-suits also trigger similar rules. 
 $L_1$ rule H2 always generate a \emph{bid} while the $L_2$ rules of model M lead to a distinction according to the number of cards in \Ss{} and the number of HCP: the more cards you have in \Ss{}, the more HCP you need to make the \emph{bid} decision. If we look more closely, Tilde before post-processing had an embryo of similar rule since it predicted a \emph{bid} with a 6-card-suit (or more) and exactly 4 HCP. The support being very small (2 examples), it had been ignored. 
\subsection{Summary}The existence of these transition zones is beyond doubt for experts, but their very identification is likely to advance knowledge in the field. New experiments conducted on these transition zones, with a more specific vocabulary devised with experts in the loop, could lead to a more accurate elicitation of the black box model.
%
\addcontentsline{toc}{section}{Appendix 1}
\addcontentsline{toc}{section}{Appendix 2}
%
%

\end{document}